# ADAM: A METHOD FOR STOCHASTIC OPTIMIZATION

**Diederik P. Kingma**[*]
University of Amsterdam, OpenAI
dpkingma@openai.com

**Jimmy Lei Ba**[*]
University of Toronto
jimmy@psi.utoronto.ca

## ABSTRACT

We introduce *Adam*, an algorithm for first-order gradient-based optimization of stochastic objective functions, based on adaptive estimates of lower-order moments. The method is straightforward to implement, is computationally efficient, has little memory requirements, is invariant to diagonal rescaling of the gradients, and is well suited for problems that are large in terms of data and/or parameters. The method is also appropriate for non-stationary objectives and problems with very noisy and/or sparse gradients. The hyper-parameters have intuitive interpretations and typically require little tuning. Some connections to related algorithms, on which *Adam* was inspired, are discussed. We also analyze the theoretical convergence properties of the algorithm and provide a regret bound on the convergence rate that is comparable to the best known results under the online convex optimization framework. Empirical results demonstrate that Adam works well in practice and compares favorably to other stochastic optimization methods. Finally, we discuss *AdaMax*, a variant of *Adam* based on the infinity norm.

## 1 INTRODUCTION

Stochastic gradient-based optimization is of core practical importance in many fields of science and engineering. Many problems in these fields can be cast as the optimization of some scalar parameterized objective function requiring maximization or minimization with respect to its parameters. If the function is differentiable w.r.t. its parameters, gradient descent is a relatively efficient optimization method, since the computation of first-order partial derivatives w.r.t. all the parameters is of the same computational complexity as just evaluating the function. Often, objective functions are stochastic. For example, many objective functions are composed of a sum of subfunctions evaluated at different subsamples of data; in this case optimization can be made more efficient by taking gradient steps w.r.t. individual subfunctions, i.e. stochastic gradient descent (SGD) or ascent. SGD proved itself as an efficient and effective optimization method that was central in many machine learning success stories, such as recent advances in deep learning (Deng et al., 2013; Krizhevsky et al., 2012; Hinton & Salakhutdinov, 2006; Hinton et al., 2012a; Graves et al., 2013). Objectives may also have other sources of noise than data subsampling, such as dropout (Hinton et al., 2012b) regularization. For all such noisy objectives, efficient stochastic optimization techniques are required. The focus of this paper is on the optimization of stochastic objectives with high-dimensional parameters spaces. In these cases, higher-order optimization methods are ill-suited, and discussion in this paper will be restricted to first-order methods.

We propose *Adam*, a method for efficient stochastic optimization that only requires first-order gradients with little memory requirement. The method computes individual adaptive learning rates for different parameters from estimates of first and second moments of the gradients; the name *Adam* is derived from adaptive moment estimation. Our method is designed to combine the advantages of two recently popular methods: AdaGrad (Duchi et al., 2011), which works well with sparse gradients, and RMSProp (Tieleman & Hinton, 2012), which works well in on-line and non-stationary settings; important connections to these and other stochastic optimization methods are clarified in section 5. Some of Adam's advantages are that the magnitudes of parameter updates are invariant to rescaling of the gradient, its stepsizes are approximately bounded by the stepsize hyperparameter, it does not require a stationary objective, it works with sparse gradients, and it naturally performs a form of step size annealing.

---

[*]Equal contribution. Author ordering determined by coin flip over a Google Hangout.

---

**Algorithm 1:** *Adam*, our proposed algorithm for stochastic optimization. See section 2 for details, and for a slightly more efficient (but less clear) order of computation. $g_t^2$ indicates the elementwise square $g_t \odot g_t$. Good default settings for the tested machine learning problems are $\alpha = 0.001$, $\beta_1 = 0.9$, $\beta_2 = 0.999$ and $\epsilon = 10^{-8}$. All operations on vectors are element-wise. With $\beta_1^t$ and $\beta_2^t$ we denote $\beta_1$ and $\beta_2$ to the power $t$.

**Require:** $\alpha$: Stepsize
**Require:** $\beta_1, \beta_2 \in [0, 1)$: Exponential decay rates for the moment estimates
**Require:** $f(\theta)$: Stochastic objective function with parameters $\theta$
**Require:** $\theta_0$: Initial parameter vector
   $m_0 \leftarrow 0$ (Initialize 1st moment vector)
   $v_0 \leftarrow 0$ (Initialize 2nd moment vector)
   $t \leftarrow 0$ (Initialize timestep)
   **while** $\theta_t$ not converged **do**
     $t \leftarrow t + 1$
     $g_t \leftarrow \nabla_\theta f_t(\theta_{t-1})$ (Get gradients w.r.t. stochastic objective at timestep $t$)
     $m_t \leftarrow \beta_1 \cdot m_{t-1} + (1 - \beta_1) \cdot g_t$ (Update biased first moment estimate)
     $v_t \leftarrow \beta_2 \cdot v_{t-1} + (1 - \beta_2) \cdot g_t^2$ (Update biased second raw moment estimate)
     $\widehat{m}_t \leftarrow m_t/(1 - \beta_1^t)$ (Compute bias-corrected first moment estimate)
     $\widehat{v}_t \leftarrow v_t/(1 - \beta_2^t)$ (Compute bias-corrected second raw moment estimate)
     $\theta_t \leftarrow \theta_{t-1} - \alpha \cdot \widehat{m}_t/(\sqrt{\widehat{v}_t} + \epsilon)$ (Update parameters)
   **end while**
   **return** $\theta_t$ (Resulting parameters)

---

In section 2 we describe the algorithm and the properties of its update rule. Section 3 explains our initialization bias correction technique, and section 4 provides a theoretical analysis of Adam's convergence in online convex programming. Empirically, our method consistently outperforms other methods for a variety of models and datasets, as shown in section 6. Overall, we show that Adam is a versatile algorithm that scales to large-scale high-dimensional machine learning problems.

## 2 ALGORITHM

See algorithm 1 for pseudo-code of our proposed algorithm *Adam*. Let $f(\theta)$ be a noisy objective function: a stochastic scalar function that is differentiable w.r.t. parameters $\theta$. We are interested in minimizing the expected value of this function, $\mathbb{E}[f(\theta)]$ w.r.t. its parameters $\theta$. With $f_1(\theta), ..., f_T(\theta)$ we denote the realisations of the stochastic function at subsequent timesteps $1, ..., T$. The stochasticity might come from the evaluation at random subsamples (minibatches) of datapoints, or arise from inherent function noise. With $g_t = \nabla_\theta f_t(\theta)$ we denote the gradient, i.e. the vector of partial derivatives of $f_t$, w.r.t $\theta$ evaluated at timestep $t$.

The algorithm updates exponential moving averages of the gradient ($m_t$) and the squared gradient ($v_t$) where the hyper-parameters $\beta_1, \beta_2 \in [0, 1)$ control the exponential decay rates of these moving averages. The moving averages themselves are estimates of the 1st moment (the mean) and the 2nd raw moment (the uncentered variance) of the gradient. However, these moving averages are initialized as (vectors of) 0's, leading to moment estimates that are biased towards zero, especially during the initial timesteps, and especially when the decay rates are small (i.e. the $\beta$s are close to 1). The good news is that this initialization bias can be easily counteracted, resulting in bias-corrected estimates $\widehat{m}_t$ and $\widehat{v}_t$. See section 3 for more details.

Note that the efficiency of algorithm 1 can, at the expense of clarity, be improved upon by changing the order of computation, e.g. by replacing the last three lines in the loop with the following lines: $\alpha_t = \alpha \cdot \sqrt{1 - \beta_2^t}/(1 - \beta_1^t)$ and $\theta_t \leftarrow \theta_{t-1} - \alpha_t \cdot m_t/(\sqrt{v_t} + \hat{\epsilon})$.

### 2.1 ADAM'S UPDATE RULE

An important property of Adam's update rule is its careful choice of stepsizes. Assuming $\epsilon = 0$, the effective step taken in parameter space at timestep $t$ is $\Delta_t = \alpha \cdot \widehat{m}_t/\sqrt{\widehat{v}_t}$. The effective stepsize has two upper bounds: $|\Delta_t| \leq \alpha \cdot (1 - \beta_1)/\sqrt{1 - \beta_2}$ in the case $(1 - \beta_1) > \sqrt{1 - \beta_2}$, and $|\Delta_t| \leq \alpha$ otherwise. The first case only happens in the most severe case of sparsity: when a gradient has been zero at all timesteps except at the current timestep. For less sparse cases, the effective stepsize will be smaller. When $(1 - \beta_1) = \sqrt{1 - \beta_2}$ we have that $|\widehat{m}_t/\sqrt{\widehat{v}_t}| < 1$ therefore $|\Delta_t| < \alpha$. In more common scenarios, we will have that $\widehat{m}_t/\sqrt{\widehat{v}_t} \approx \pm 1$ since $|\mathbb{E}[g]/\sqrt{\mathbb{E}[g^2]}| \leq 1$. The effective magnitude of the steps taken in parameter space at each timestep are approximately bounded by the stepsize setting $\alpha$, i.e., $|\Delta_t| \lessapprox \alpha$. This can be understood as establishing a *trust region* around the current parameter value, beyond which the current gradient estimate does not provide sufficient information. This typically makes it relatively easy to know the right scale of $\alpha$ in advance. For many machine learning models, for instance, we often know in advance that good optima are with high probability within some set region in parameter space; it is not uncommon, for example, to have a prior distribution over the parameters. Since $\alpha$ sets (an upper bound of) the magnitude of steps in parameter space, we can often deduce the right order of magnitude of $\alpha$ such that optima can be reached from $\theta_0$ within some number of iterations. With a slight abuse of terminology, we will call the ratio $\widehat{m}_t/\sqrt{\widehat{v}_t}$ the *signal-to-noise* ratio ($SNR$). With a smaller SNR the effective stepsize $\Delta_t$ will be closer to zero. This is a desirable property, since a smaller SNR means that there is greater uncertainty about whether the direction of $\widehat{m}_t$ corresponds to the direction of the true gradient. For example, the SNR value typically becomes closer to 0 towards an optimum, leading to smaller effective steps in parameter space: a form of automatic annealing. The effective stepsize $\Delta_t$ is also invariant to the scale of the gradients; rescaling the gradients $g$ with factor $c$ will scale $\widehat{m}_t$ with a factor $c$ and $\widehat{v}_t$ with a factor $c^2$, which cancel out: $(c \cdot \widehat{m}_t)/(\sqrt{c^2 \cdot \widehat{v}_t}) = \widehat{m}_t/\sqrt{\widehat{v}_t}$.

## 3 INITIALIZATION BIAS CORRECTION

As explained in section 2, Adam utilizes initialization bias correction terms. We will here derive the term for the second moment estimate; the derivation for the first moment estimate is completely analogous. Let $g$ be the gradient of the stochastic objective $f$, and we wish to estimate its second raw moment (uncentered variance) using an exponential moving average of the squared gradient, with decay rate $\beta_2$. Let $g_1, ..., g_T$ be the gradients at subsequent timesteps, each a draw from an underlying gradient distribution $g_t \sim p(g_t)$. Let us initialize the exponential moving average as $v_0 = 0$ (a vector of zeros). First note that the update at timestep $t$ of the exponential moving average $v_t = \beta_2 \cdot v_{t-1} + (1 - \beta_2) \cdot g_t^2$ (where $g_t^2$ indicates the elementwise square $g_t \odot g_t$) can be written as a function of the gradients at all previous timesteps:

$$v_t = (1 - \beta_2) \sum_{i=1}^{t} \beta_2^{t-i} \cdot g_i^2 \qquad (1)$$

We wish to know how $\mathbb{E}[v_t]$, the expected value of the exponential moving average at timestep $t$, relates to the true second moment $\mathbb{E}[g_t^2]$, so we can correct for the discrepancy between the two. Taking expectations of the left-hand and right-hand sides of eq. (1):

$$\mathbb{E}[v_t] = \mathbb{E}\left[(1 - \beta_2) \sum_{i=1}^{t} \beta_2^{t-i} \cdot g_i^2\right] \qquad (2)$$

$$= \mathbb{E}[g_t^2] \cdot (1 - \beta_2) \sum_{i=1}^{t} \beta_2^{t-i} + \zeta \qquad (3)$$

$$= \mathbb{E}[g_t^2] \cdot (1 - \beta_2^t) + \zeta \qquad (4)$$

where $\zeta = 0$ if the true second moment $\mathbb{E}[g_i^2]$ is stationary; otherwise $\zeta$ can be kept small since the exponential decay rate $\beta_1$ can (and should) be chosen such that the exponential moving average assigns small weights to gradients too far in the past. What is left is the term $(1 - \beta_2^t)$ which is caused by initializing the running average with zeros. In algorithm 1 we therefore divide by this term to correct the initialization bias.

In case of sparse gradients, for a reliable estimate of the second moment one needs to average over many gradients by chosing a small value of $\beta_2$; however it is exactly this case of small $\beta_2$ where a lack of initialisation bias correction would lead to initial steps that are much larger.

## 4 CONVERGENCE ANALYSIS

We analyze the convergence of Adam using the online learning framework proposed in (Zinkevich, 2003). Given an arbitrary, unknown sequence of convex cost functions $f_1(\theta), f_2(\theta),..., f_T(\theta)$. At each time $t$, our goal is to predict the parameter $\theta_t$ and evaluate it on a previously unknown cost function $f_t$. Since the nature of the sequence is unknown in advance, we evaluate our algorithm using the regret, that is the sum of all the previous difference between the online prediction $f_t(\theta_t)$ and the best fixed point parameter $f_t(\theta^*)$ from a feasible set $\mathcal{X}$ for all the previous steps. Concretely, the regret is defined as:

$$R(T) = \sum_{t=1}^{T}[f_t(\theta_t) - f_t(\theta^*)] \quad (5)$$

where $\theta^* = \arg\min_{\theta \in \mathcal{X}} \sum_{t=1}^{T} f_t(\theta)$. We show Adam has $O(\sqrt{T})$ regret bound and a proof is given in the appendix. Our result is comparable to the best known bound for this general convex online learning problem. We also use some definitions simplify our notation, where $g_t \triangleq \nabla f_t(\theta_t)$ and $g_{t,i}$ as the $i^{\text{th}}$ element. We define $g_{1:t,i} \in \mathbb{R}^t$ as a vector that contains the $i^{\text{th}}$ dimension of the gradients over all iterations till $t$, $g_{1:t,i} = [g_{1,i}, g_{2,i}, \cdots, g_{t,i}]$. Also, we define $\gamma \triangleq \frac{\beta_1^2}{\sqrt{\beta_2}}$. Our following theorem holds when the learning rate $\alpha_t$ is decaying at a rate of $t^{-\frac{1}{2}}$ and first moment running average coefficient $\beta_{1,t}$ decay exponentially with $\lambda$, that is typically close to 1, e.g. $1 - 10^{-8}$.

**Theorem 4.1.** *Assume that the function $f_t$ has bounded gradients, $\|\nabla f_t(\theta)\|_2 \leq G$, $\|\nabla f_t(\theta)\|_\infty \leq G_\infty$ for all $\theta \in R^d$ and distance between any $\theta_t$ generated by Adam is bounded, $\|\theta_n - \theta_m\|_2 \leq D$, $\|\theta_m - \theta_n\|_\infty \leq D_\infty$ for any $m, n \in \{1, ..., T\}$, and $\beta_1, \beta_2 \in [0, 1)$ satisfy $\frac{\beta_1^2}{\sqrt{\beta_2}} < 1$. Let $\alpha_t = \frac{\alpha}{\sqrt{t}}$ and $\beta_{1,t} = \beta_1 \lambda^{t-1}, \lambda \in (0, 1)$. Adam achieves the following guarantee, for all $T \geq 1$.*

$$R(T) \leq \frac{D^2}{2\alpha(1-\beta_1)}\sum_{i=1}^{d}\sqrt{T\hat{v}_{T,i}} + \frac{\alpha(1+\beta_1)G_\infty}{(1-\beta_1)\sqrt{1-\beta_2}(1-\gamma)^2}\sum_{i=1}^{d}\|g_{1:T,i}\|_2 + \sum_{i=1}^{d}\frac{D_\infty^2 G_\infty \sqrt{1-\beta_2}}{2\alpha(1-\beta_1)(1-\lambda)^2}$$

Our Theorem 4.1 implies when the data features are sparse and bounded gradients, the summation term can be much smaller than its upper bound $\sum_{i=1}^{d}\|g_{1:T,i}\|_2 << dG_\infty\sqrt{T}$ and $\sum_{i=1}^{d}\sqrt{T\hat{v}_{T,i}} << dG_\infty\sqrt{T}$, in particular if the class of function and data features are in the form of section 1.2 in (Duchi et al., 2011). Their results for the expected value $\mathbb{E}[\sum_{i=1}^{d}\|g_{1:T,i}\|_2]$ also apply to Adam. In particular, the adaptive method, such as Adam and Adagrad, can achieve $O(\log d\sqrt{T})$, an improvement over $O(\sqrt{dT})$ for the non-adaptive method. Decaying $\beta_{1,t}$ towards zero is important in our theoretical analysis and also matches previous empirical findings, e.g. (Sutskever et al., 2013) suggests reducing the momentum coefficient in the end of training can improve convergence.

Finally, we can show the average regret of Adam converges,

**Corollary 4.2.** *Assume that the function $f_t$ has bounded gradients, $\|\nabla f_t(\theta)\|_2 \leq G$, $\|\nabla f_t(\theta)\|_\infty \leq G_\infty$ for all $\theta \in R^d$ and distance between any $\theta_t$ generated by Adam is bounded, $\|\theta_n - \theta_m\|_2 \leq D$, $\|\theta_m - \theta_n\|_\infty \leq D_\infty$ for any $m, n \in \{1, ..., T\}$. Adam achieves the following guarantee, for all $T \geq 1$.*

$$\frac{R(T)}{T} = O(\frac{1}{\sqrt{T}})$$

This result can be obtained by using Theorem 4.1 and $\sum_{i=1}^{d}\|g_{1:T,i}\|_2 \leq dG_\infty\sqrt{T}$. Thus, $\lim_{T\to\infty}\frac{R(T)}{T} = 0$.

## 5 RELATED WORK

Optimization methods bearing a direct relation to Adam are RMSProp (Tieleman & Hinton, 2012; Graves, 2013) and AdaGrad (Duchi et al., 2011); these relationships are discussed below. Other stochastic optimization methods include vSGD (Schaul et al., 2012), AdaDelta (Zeiler, 2012) and the natural Newton method from Roux & Fitzgibbon (2010), all setting stepsizes by estimating curvature from first-order information. The Sum-of-Functions Optimizer (SFO) (Sohl-Dickstein et al., 2014) is a quasi-Newton method based on minibatches, but (unlike Adam) has memory requirements linear in the number of minibatch partitions of a dataset, which is often infeasible on memory-constrained systems such as a GPU. Like natural gradient descent (NGD) (Amari, 1998), Adam employs a preconditioner that adapts to the geometry of the data, since $\widehat{v}_t$ is an approximation to the diagonal of the Fisher information matrix (Pascanu & Bengio, 2013); however, Adam's preconditioner (like AdaGrad's) is more conservative in its adaption than vanilla NGD by preconditioning with the square root of the inverse of the diagonal Fisher information matrix approximation.

**RMSProp:** An optimization method closely related to Adam is RMSProp (Tieleman & Hinton, 2012). A version with momentum has sometimes been used (Graves, 2013). There are a few important differences between RMSProp with momentum and Adam: RMSProp with momentum generates its parameter updates using a momentum on the rescaled gradient, whereas Adam updates are directly estimated using a running average of first and second moment of the gradient. RMSProp also lacks a bias-correction term; this matters most in case of a value of $\beta_2$ close to 1 (required in case of sparse gradients), since in that case not correcting the bias leads to very large stepsizes and often divergence, as we also empirically demonstrate in section 6.4.

**AdaGrad:** An algorithm that works well for sparse gradients is AdaGrad (Duchi et al., 2011). Its basic version updates parameters as $\theta_{t+1} = \theta_t - \alpha \cdot g_t / \sqrt{\sum_{i=1}^{t} g_i^2}$. Note that if we choose $\beta_2$ to be infinitesimally close to 1 from below, then $\lim_{\beta_2 \to 1} \widehat{v}_t = t^{-1} \cdot \sum_{i=1}^{t} g_i^2$. AdaGrad corresponds to a version of Adam with $\beta_1 = 0$, infinitesimal $(1 - \beta_2)$ and a replacement of $\alpha$ by an annealed version $\alpha_t = \alpha \cdot t^{-1/2}$, namely $\theta_t - \alpha \cdot t^{-1/2} \cdot \widehat{m}_t / \sqrt{\lim_{\beta_2 \to 1} \widehat{v}_t} = \theta_t - \alpha \cdot t^{-1/2} \cdot g_t / \sqrt{t^{-1} \cdot \sum_{i=1}^{t} g_i^2} = \theta_t - \alpha \cdot g_t / \sqrt{\sum_{i=1}^{t} g_i^2}$. Note that this direct correspondence between Adam and Adagrad does not hold when removing the bias-correction terms; without bias correction, like in RMSProp, a $\beta_2$ infinitesimally close to 1 would lead to infinitely large bias, and infinitely large parameter updates.

## 6 EXPERIMENTS

To empirically evaluate the proposed method, we investigated different popular machine learning models, including logistic regression, multilayer fully connected neural networks and deep convolutional neural networks. Using large models and datasets, we demonstrate Adam can efficiently solve practical deep learning problems.

We use the same parameter initialization when comparing different optimization algorithms. The hyper-parameters, such as learning rate and momentum, are searched over a dense grid and the results are reported using the best hyper-parameter setting.

### 6.1 EXPERIMENT: LOGISTIC REGRESSION

We evaluate our proposed method on L2-regularized multi-class logistic regression using the MNIST dataset. Logistic regression has a well-studied convex objective, making it suitable for comparison of different optimizers without worrying about local minimum issues. The stepsize $\alpha$ in our logistic regression experiments is adjusted by $1/\sqrt{t}$ decay, namely $\alpha_t = \frac{\alpha}{\sqrt{t}}$ that matches with our theoratical prediction from section 4. The logistic regression classifies the class label directly on the 784 dimension image vectors. We compare Adam to accelerated SGD with Nesterov momentum and Adagrad using minibatch size of 128. According to Figure 1, we found that the Adam yields similar convergence as SGD with momentum and both converge faster than Adagrad.

As discussed in (Duchi et al., 2011), Adagrad can efficiently deal with sparse features and gradients as one of its main theoretical results whereas SGD is low at learning rare features. Adam with $1/\sqrt{t}$ decay on its stepsize should theoratically match the performance of Adagrad. We examine the sparse feature problem using IMDB movie review dataset from (Maas et al., 2011). We pre-process the IMDB movie reviews into bag-of-words (BoW) feature vectors including the first 10,000 most frequent words. The 10,000 dimension BoW feature vector for each review is highly sparse. As suggested in (Wang & Manning, 2013), 50% dropout noise can be applied to the BoW features during training to prevent over-fitting. In figure 1, Adagrad outperforms SGD with Nesterov momentum by a large margin both with and without dropout noise. Adam converges as fast as Adagrad. The empirical performance of Adam is consistent with our theoretical findings in sections 2 and 4. Similar to Adagrad, Adam can take advantage of sparse features and obtain faster convergence rate than normal SGD with momentum.

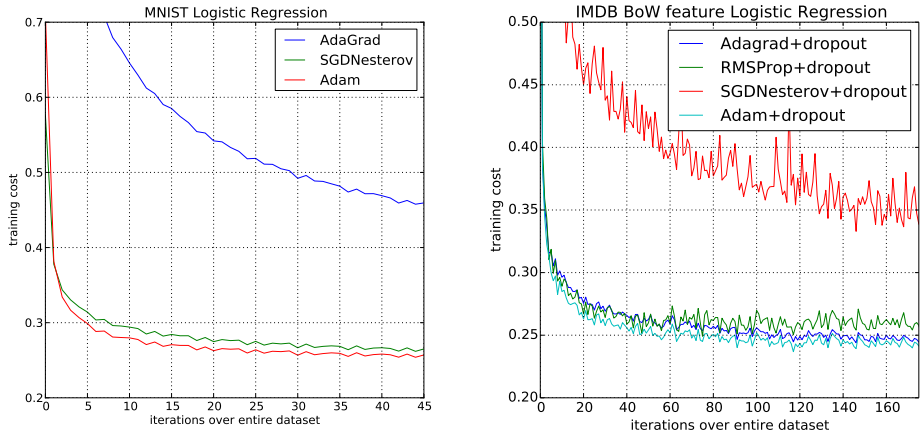

Figure 1: Logistic regression training negative log likelihood on MNIST images and IMDB movie reviews with 10,000 bag-of-words (BoW) feature vectors.

### 6.2 EXPERIMENT: MULTI-LAYER NEURAL NETWORKS

Multi-layer neural network are powerful models with non-convex objective functions. Although our convergence analysis does not apply to non-convex problems, we empirically found that Adam often outperforms other methods in such cases. In our experiments, we made model choices that are consistent with previous publications in the area; a neural network model with two fully connected hidden layers with 1000 hidden units each and ReLU activation are used for this experiment with minibatch size of 128.

First, we study different optimizers using the standard deterministic cross-entropy objective function with $L_2$ weight decay on the parameters to prevent over-fitting. The sum-of-functions (SFO) method (Sohl-Dickstein et al., 2014) is a recently proposed quasi-Newton method that works with minibatches of data and has shown good performance on optimization of multi-layer neural networks. We used their implementation and compared with Adam to train such models. Figure 2 shows that Adam makes faster progress in terms of both the number of iterations and wall-clock time. Due to the cost of updating curvature information, SFO is 5-10x slower per iteration compared to Adam, and has a memory requirement that is linear in the number minibatches.

Stochastic regularization methods, such as dropout, are an effective way to prevent over-fitting and often used in practice due to their simplicity. SFO assumes deterministic subfunctions, and indeed failed to converge on cost functions with stochastic regularization. We compare the effectiveness of Adam to other stochastic first order methods on multi-layer neural networks trained with dropout noise. Figure 2 shows our results; Adam shows better convergence than other methods.

### 6.3 EXPERIMENT: CONVOLUTIONAL NEURAL NETWORKS

Convolutional neural networks (CNNs) with several layers of convolution, pooling and non-linear units have shown considerable success in computer vision tasks. Unlike most fully connected neural nets, weight sharing in CNNs results in vastly different gradients in different layers. A smaller learning rate for the convolution layers is often used in practice when applying SGD. We show the effectiveness of Adam in deep CNNs. Our CNN architecture has three alternating stages of 5x5 convolution filters and 3x3 max pooling with stride of 2 that are followed by a fully connected layer of 1000 rectified linear hidden units (ReLU's). The input image are pre-processed by whitening, and dropout noise is applied to the input layer and fully connected layer. The minibatch size is also set to 128 similar to previous experiments.

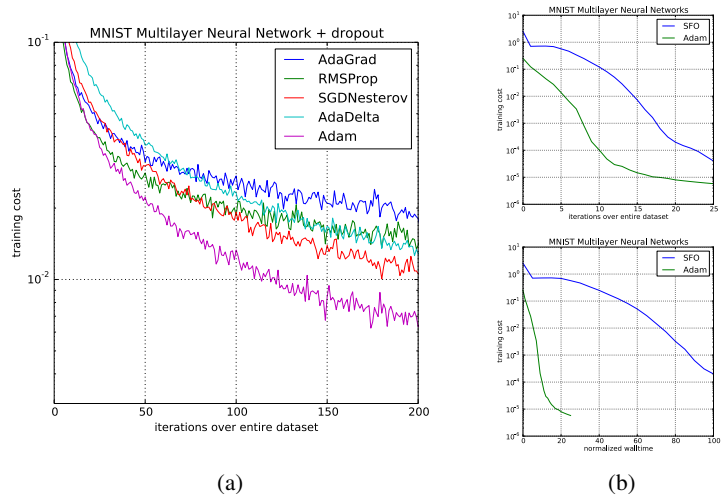

Figure 2: Training of multilayer neural networks on MNIST images. (a) Neural networks using dropout stochastic regularization. (b) Neural networks with deterministic cost function. We compare with the sum-of-functions (SFO) optimizer (Sohl-Dickstein et al., 2014)

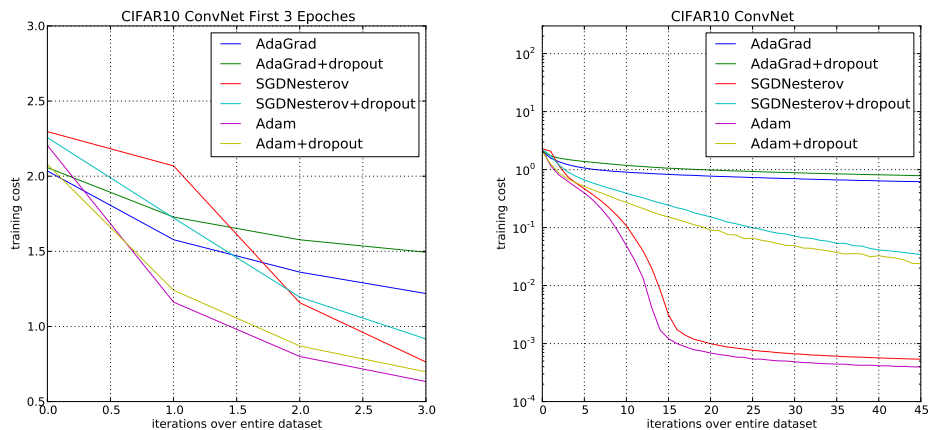

Figure 3: Convolutional neural networks training cost. (left) Training cost for the first three epochs. (right) Training cost over 45 epochs. CIFAR-10 with c64-c64-c128-1000 architecture.

Interestingly, although both Adam and Adagrad make rapid progress lowering the cost in the initial stage of the training, shown in Figure 3 (left), Adam and SGD eventually converge considerably faster than Adagrad for CNNs shown in Figure 3 (right). We notice the second moment estimate $\widehat{v}_t$ vanishes to zeros after a few epochs and is dominated by the $\epsilon$ in algorithm 1. The second moment estimate is therefore a poor approximation to the geometry of the cost function in CNNs comparing to fully connected network from Section 6.2. Whereas, reducing the minibatch variance through the first moment is more important in CNNs and contributes to the speed-up. As a result, Adagrad converges much slower than others in this particular experiment. Though Adam shows marginal improvement over SGD with momentum, it adapts learning rate scale for different layers instead of hand picking manually as in SGD.

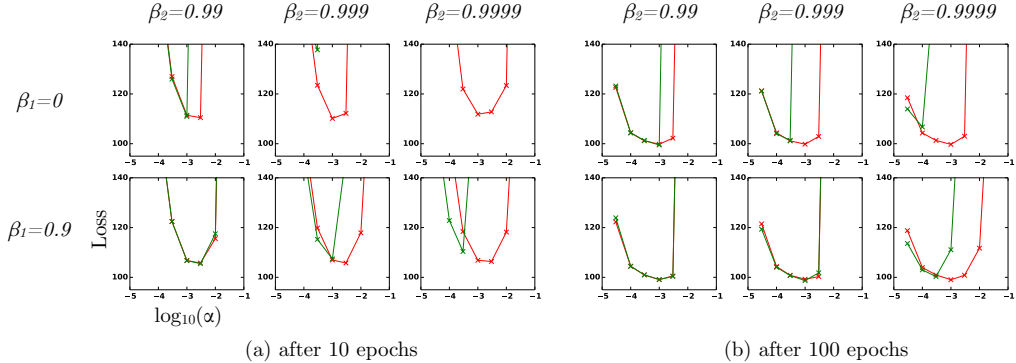

Figure 4: Effect of bias-correction terms (red line) versus no bias correction terms (green line) after 10 epochs (left) and 100 epochs (right) on the loss (y-axes) when learning a Variational Auto-Encoder (VAE) (Kingma & Welling, 2013), for different settings of stepsize $\alpha$ (x-axes) and hyper-parameters $\beta_1$ and $\beta_2$.

### 6.4 EXPERIMENT: BIAS-CORRECTION TERM

We also empirically evaluate the effect of the bias correction terms explained in sections 2 and 3. Discussed in section 5, removal of the bias correction terms results in a version of RMSProp (Tieleman & Hinton, 2012) with momentum. We vary the $\beta_1$ and $\beta_2$ when training a variational auto-encoder (VAE) with the same architecture as in (Kingma & Welling, 2013) with a single hidden layer with 500 hidden units with softplus nonlinearities and a 50-dimensional spherical Gaussian latent variable. We iterated over a broad range of hyper-parameter choices, i.e. $\beta_1 \in [0, 0.9]$ and $\beta_2 \in [0.99, 0.999, 0.9999]$, and $\log_{10}(\alpha) \in [-5, ..., -1]$. Values of $\beta_2$ close to 1, required for robustness to sparse gradients, results in larger initialization bias; therefore we expect the bias correction term is important in such cases of slow decay, preventing an adverse effect on optimization.

In Figure 4, values $\beta_2$ close to 1 indeed lead to instabilities in training when no bias correction term was present, especially at first few epochs of the training. The best results were achieved with small values of $(1 - \beta_2)$ and bias correction; this was more apparent towards the end of optimization when gradients tends to become sparser as hidden units specialize to specific patterns. In summary, Adam performed equal or better than RMSProp, regardless of hyper-parameter setting.

## 7 EXTENSIONS

### 7.1 ADAMAX

In Adam, the update rule for individual weights is to scale their gradients inversely proportional to a (scaled) $L^2$ norm of their individual current and past gradients. We can generalize the $L^2$ norm based update rule to a $L^p$ norm based update rule. Such variants become numerically unstable for large $p$. However, in the special case where we let $p \to \infty$, a surprisingly simple and stable algorithm emerges; see algorithm 2. We'll now derive the algorithm. Let, in case of the $L^p$ norm, the stepsize at time $t$ be inversely proportional to $v_t^{1/p}$, where:

$$v_t = \beta_2^p v_{t-1} + (1 - \beta_2^p)|g_t|^p \qquad (6)$$

$$= (1 - \beta_2^p) \sum_{i=1}^{t} \beta_2^{p(t-i)} \cdot |g_i|^p \qquad (7)$$

**Algorithm 2:** *AdaMax*, a variant of Adam based on the infinity norm. See section 7.1 for details. Good default settings for the tested machine learning problems are $\alpha = 0.002$, $\beta_1 = 0.9$ and $\beta_2 = 0.999$. With $\beta_1^t$ we denote $\beta_1$ to the power $t$. Here, $(\alpha/(1 - \beta_1^t))$ is the learning rate with the bias-correction term for the first moment. All operations on vectors are element-wise.

**Require:** $\alpha$: Stepsize
**Require:** $\beta_1, \beta_2 \in [0, 1)$: Exponential decay rates
**Require:** $f(\theta)$: Stochastic objective function with parameters $\theta$
**Require:** $\theta_0$: Initial parameter vector
    $m_0 \leftarrow 0$ (Initialize 1st moment vector)
    $u_0 \leftarrow 0$ (Initialize the exponentially weighted infinity norm)
    $t \leftarrow 0$ (Initialize timestep)
    **while** $\theta_t$ not converged **do**
        $t \leftarrow t + 1$
        $g_t \leftarrow \nabla_\theta f_t(\theta_{t-1})$ (Get gradients w.r.t. stochastic objective at timestep $t$)
        $m_t \leftarrow \beta_1 \cdot m_{t-1} + (1 - \beta_1) \cdot g_t$ (Update biased first moment estimate)
        $u_t \leftarrow \max(\beta_2 \cdot u_{t-1}, |g_t|)$ (Update the exponentially weighted infinity norm)
        $\theta_t \leftarrow \theta_{t-1} - (\alpha/(1 - \beta_1^t)) \cdot m_t/u_t$ (Update parameters)
    **end while**
    **return** $\theta_t$ (Resulting parameters)

Note that the decay term is here equivalently parameterised as $\beta_2^p$ instead of $\beta_2$. Now let $p \to \infty$, and define $u_t = \lim_{p \to \infty}(v_t)^{1/p}$, then:

$$u_t = \lim_{p \to \infty}(v_t)^{1/p} = \lim_{p \to \infty}\left((1 - \beta_2^p)\sum_{i=1}^{t}\beta_2^{p(t-i)} \cdot |g_i|^p\right)^{1/p} \tag{8}$$

$$= \lim_{p \to \infty}(1 - \beta_2^p)^{1/p}\left(\sum_{i=1}^{t}\beta_2^{p(t-i)} \cdot |g_i|^p\right)^{1/p} \tag{9}$$

$$= \lim_{p \to \infty}\left(\sum_{i=1}^{t}\left(\beta_2^{(t-i)} \cdot |g_i|\right)^p\right)^{1/p} \tag{10}$$

$$= \max\left(\beta_2^{t-1}|g_1|, \beta_2^{t-2}|g_2|, \ldots, \beta_2|g_{t-1}|, |g_t|\right) \tag{11}$$

Which corresponds to the remarkably simple recursive formula:

$$u_t = \max(\beta_2 \cdot u_{t-1}, |g_t|) \tag{12}$$

with initial value $u_0 = 0$. Note that, conveniently enough, we don't need to correct for initialization bias in this case. Also note that the magnitude of parameter updates has a simpler bound with AdaMax than Adam, namely: $|\Delta_t| \leq \alpha$.

### 7.2 TEMPORAL AVERAGING

Since the last iterate is noisy due to stochastic approximation, better generalization performance is often achieved by averaging. Previously in Moulines & Bach (2011), Polyak-Ruppert averaging (Polyak & Juditsky, 1992; Ruppert, 1988) has been shown to improve the convergence of standard SGD, where $\bar{\theta}_t = \frac{1}{t}\sum_{k=1}^{n}\theta_k$. Alternatively, an exponential moving average over the parameters can be used, giving higher weight to more recent parameter values. This can be trivially implemented by adding one line to the inner loop of algorithms 1 and 2: $\bar{\theta}_t \leftarrow \beta_2 \cdot \bar{\theta}_{t-1} + (1 - \beta_2)\theta_t$, with $\bar{\theta}_0 = 0$. Initalization bias can again be corrected by the estimator $\widehat{\bar{\theta}_t} = \bar{\theta}_t/(1 - \beta_2^t)$.

## 8 CONCLUSION

We have introduced a simple and computationally efficient algorithm for gradient-based optimization of stochastic objective functions. Our method is aimed towards machine learning problems with large datasets and/or high-dimensional parameter spaces. The method combines the advantages of two recently popular optimization methods: the ability of AdaGrad to deal with sparse gradients, and the ability of RMSProp to deal with non-stationary objectives. The method is straightforward to implement and requires little memory. The experiments confirm the analysis on the rate of convergence in convex problems. Overall, we found Adam to be robust and well-suited to a wide range of non-convex optimization problems in the field machine learning.

## 9 ACKNOWLEDGMENTS

This paper would probably not have existed without the support of Google Deepmind. We would like to give special thanks to Ivo Danihelka, and Tom Schaul for coining the name Adam. Thanks to Kai Fan from Duke University for spotting an error in the original AdaMax derivation. Experiments in this work were partly carried out on the Dutch national e-infrastructure with the support of SURF Foundation. Diederik Kingma is supported by the Google European Doctorate Fellowship in Deep Learning.

## 10 APPENDIX

### 10.1 CONVERGENCE PROOF

**Definition 10.1.** *A function $f : R^d \to R$ is convex if for all $x, y \in R^d$, for all $\lambda \in [0, 1]$,*
$$\lambda f(x) + (1 - \lambda)f(y) \geq f(\lambda x + (1 - \lambda)y)$$

Also, notice that a convex function can be lower bounded by a hyperplane at its tangent.

**Lemma 10.2.** *If a function $f : R^d \to R$ is convex, then for all $x, y \in R^d$,*
$$f(y) \geq f(x) + \nabla f(x)^T (y - x)$$

The above lemma can be used to upper bound the regret and our proof for the main theorem is constructed by substituting the hyperplane with the Adam update rules.

The following two lemmas are used to support our main theorem. We also use some definitions simplify our notation, where $g_t \triangleq \nabla f_t(\theta_t)$ and $g_{t,i}$ as the $i^{\text{th}}$ element. We define $g_{1:t,i} \in \mathbb{R}^t$ as a vector that contains the $i^{\text{th}}$ dimension of the gradients over all iterations till $t$, $g_{1:t,i} = [g_{1,i}, g_{2,i}, \cdots, g_{t,i}]$

**Lemma 10.3.** *Let $g_t = \nabla f_t(\theta_t)$ and $g_{1:t}$ be defined as above and bounded, $\|g_t\|_2 \leq G$, $\|g_t\|_\infty \leq G_\infty$. Then,*
$$\sum_{t=1}^{T} \sqrt{\frac{g_{t,i}^2}{t}} \leq 2G_\infty \|g_{1:T,i}\|_2$$

*Proof.* We will prove the inequality using induction over T.

The base case for $T = 1$, we have $\sqrt{g_{1,i}^2} \leq 2G_\infty \|g_{1,i}\|_2$.

For the inductive step,
$$\sum_{t=1}^{T} \sqrt{\frac{g_{t,i}^2}{t}} = \sum_{t=1}^{T-1} \sqrt{\frac{g_{t,i}^2}{t}} + \sqrt{\frac{g_{T,i}^2}{T}}$$
$$\leq 2G_\infty \|g_{1:T-1,i}\|_2 + \sqrt{\frac{g_{T,i}^2}{T}}$$
$$= 2G_\infty \sqrt{\|g_{1:T,i}\|_2^2 - g_T^2} + \sqrt{\frac{g_{T,i}^2}{T}}$$

From, $\|g_{1:T,i}\|_2^2 - g_{T,i}^2 + \frac{g_{T,i}^4}{4\|g_{1:T,i}\|_2^2} \geq \|g_{1:T,i}\|_2^2 - g_{T,i}^2$, we can take square root of both side and have,
$$\sqrt{\|g_{1:T,i}\|_2^2 - g_{T,i}^2} \leq \|g_{1:T,i}\|_2 - \frac{g_{T,i}^2}{2\|g_{1:T,i}\|_2}$$
$$\leq \|g_{1:T,i}\|_2 - \frac{g_{T,i}^2}{2\sqrt{TG_\infty^2}}$$

Rearrange the inequality and substitute the $\sqrt{\|g_{1:T,i}\|_2^2 - g_{T,i}^2}$ term,
$$G_\infty \sqrt{\|g_{1:T,i}\|_2^2 - g_T^2} + \sqrt{\frac{g_{T,i}^2}{T}} \leq 2G_\infty \|g_{1:T,i}\|_2$$

□

**Lemma 10.4.** *Let $\gamma \triangleq \frac{\beta_1^2}{\sqrt{\beta_2}}$. For $\beta_1, \beta_2 \in [0, 1)$ that satisfy $\frac{\beta_1^2}{\sqrt{\beta_2}} < 1$ and bounded $g_t$, $\|g_t\|_2 \leq G$, $\|g_t\|_\infty \leq G_\infty$, the following inequality holds*

$$\sum_{t=1}^{T} \frac{\widehat{m}_{t,i}^2}{\sqrt{t\widehat{v}_{t,i}}} \leq \frac{2}{1-\gamma} \frac{1}{\sqrt{1-\beta_2}} \|g_{1:T,i}\|_2$$

*Proof.* Under the assumption, $\frac{\sqrt{1-\beta_2^t}}{(1-\beta_1^t)^2} \leq \frac{1}{(1-\beta_1)^2}$. We can expand the last term in the summation using the update rules in Algorithm 1,

$$\sum_{t=1}^{T} \frac{\widehat{m}_{t,i}^2}{\sqrt{t\widehat{v}_{t,i}}} = \sum_{t=1}^{T-1} \frac{\widehat{m}_{t,i}^2}{\sqrt{t\widehat{v}_{t,i}}} + \frac{\sqrt{1-\beta_2^T}}{(1-\beta_1^T)^2} \frac{(\sum_{k=1}^{T}(1-\beta_1)\beta_1^{T-k} g_{k,i})^2}{\sqrt{T \sum_{j=1}^{T}(1-\beta_2)\beta_2^{T-j} g_{j,i}^2}}$$

$$\leq \sum_{t=1}^{T-1} \frac{\widehat{m}_{t,i}^2}{\sqrt{t\widehat{v}_{t,i}}} + \frac{\sqrt{1-\beta_2^T}}{(1-\beta_1^T)^2} \sum_{k=1}^{T} \frac{T((1-\beta_1)\beta_1^{T-k} g_{k,i})^2}{\sqrt{T \sum_{j=1}^{T}(1-\beta_2)\beta_2^{T-j} g_{j,i}^2}}$$

$$\leq \sum_{t=1}^{T-1} \frac{\widehat{m}_{t,i}^2}{\sqrt{t\widehat{v}_{t,i}}} + \frac{\sqrt{1-\beta_2^T}}{(1-\beta_1^T)^2} \sum_{k=1}^{T} \frac{T((1-\beta_1)\beta_1^{T-k} g_{k,i})^2}{\sqrt{T(1-\beta_2)\beta_2^{T-k} g_{k,i}^2}}$$

$$\leq \sum_{t=1}^{T-1} \frac{\widehat{m}_{t,i}^2}{\sqrt{t\widehat{v}_{t,i}}} + \frac{\sqrt{1-\beta_2^T}}{(1-\beta_1^T)^2} \frac{(1-\beta_1)^2}{\sqrt{T(1-\beta_2)}} \sum_{k=1}^{T} T \left(\frac{\beta_1^2}{\sqrt{\beta_2}}\right)^{T-k} \|g_{k,i}\|_2$$

$$\leq \sum_{t=1}^{T-1} \frac{\widehat{m}_{t,i}^2}{\sqrt{t\widehat{v}_{t,i}}} + \frac{T}{\sqrt{T(1-\beta_2)}} \sum_{k=1}^{T} \gamma^{T-k} \|g_{k,i}\|_2$$

Similarly, we can upper bound the rest of the terms in the summation.

$$\sum_{t=1}^{T} \frac{\widehat{m}_{t,i}^2}{\sqrt{t\widehat{v}_{t,i}}} \leq \sum_{t=1}^{T} \frac{\|g_{t,i}\|_2}{\sqrt{t(1-\beta_2)}} \sum_{j=0}^{T-t} t\gamma^j$$

$$\leq \sum_{t=1}^{T} \frac{\|g_{t,i}\|_2}{\sqrt{t(1-\beta_2)}} \sum_{j=0}^{T} t\gamma^j$$

For $\gamma < 1$, using the upper bound on the arithmetic-geometric series, $\sum_t t\gamma^t < \frac{1}{(1-\gamma)^2}$:

$$\sum_{t=1}^{T} \frac{\|g_{t,i}\|_2}{\sqrt{t(1-\beta_2)}} \sum_{j=0}^{T} t\gamma^j \leq \frac{1}{(1-\gamma)^2 \sqrt{1-\beta_2}} \sum_{t=1}^{T} \frac{\|g_{t,i}\|_2}{\sqrt{t}}$$

Apply Lemma 10.3,

$$\sum_{t=1}^{T} \frac{\widehat{m}_{t,i}^2}{\sqrt{t\widehat{v}_{t,i}}} \leq \frac{2G_\infty}{(1-\gamma)^2 \sqrt{1-\beta_2}} \|g_{1:T,i}\|_2$$

□

To simplify the notation, we define $\gamma \triangleq \frac{\beta_1^2}{\sqrt{\beta_2}}$. Intuitively, our following theorem holds when the learning rate $\alpha_t$ is decaying at a rate of $t^{-\frac{1}{2}}$ and first moment running average coefficient $\beta_{1,t}$ decay exponentially with $\lambda$, that is typically close to 1, e.g. $1 - 10^{-8}$.

**Theorem 10.5.** *Assume that the function $f_t$ has bounded gradients, $\|\nabla f_t(\theta)\|_2 \leq G$, $\|\nabla f_t(\theta)\|_\infty \leq G_\infty$ for all $\theta \in R^d$ and distance between any $\theta_t$ generated by Adam is bounded, $\|\theta_n - \theta_m\|_2 \leq D$,*

$\|\theta_m - \theta_n\|_\infty \leq D_\infty$ *for any* $m, n \in \{1, ..., T\}$, *and* $\beta_1, \beta_2 \in [0, 1)$ *satisfy* $\frac{\beta_1^2}{\sqrt{\beta_2}} < 1$. *Let* $\alpha_t = \frac{\alpha}{\sqrt{t}}$ *and* $\beta_{1,t} = \beta_1 \lambda^{t-1}, \lambda \in (0, 1)$. *Adam achieves the following guarantee, for all* $T \geq 1$.

$$R(T) \leq \frac{D^2}{2\alpha(1-\beta_1)} \sum_{i=1}^d \sqrt{T\widehat{v}_{T,i}} + \frac{\alpha(\beta_1+1)G_\infty}{(1-\beta_1)\sqrt{1-\beta_2}(1-\gamma)^2} \sum_{i=1}^d \|g_{1:T,i}\|_2 + \sum_{i=1}^d \frac{D_\infty^2 G_\infty \sqrt{1-\beta_2}}{2\alpha(1-\beta_1)(1-\lambda)^2}$$

*Proof.* Using Lemma 10.2, we have,

$$f_t(\theta_t) - f_t(\theta^*) \leq g_t^T (\theta_t - \theta^*) = \sum_{i=1}^d g_{t,i}(\theta_{t,i} - \theta^*_{,i})$$

From the update rules presented in algorithm 1,

$$\theta_{t+1} = \theta_t - \alpha_t \widehat{m}_t / \sqrt{\widehat{v}_t}$$
$$= \theta_t - \frac{\alpha_t}{1 - \beta_1^t} \left( \frac{\beta_{1,t}}{\sqrt{\widehat{v}_t}} m_{t-1} + \frac{(1-\beta_{1,t})}{\sqrt{\widehat{v}_t}} g_t \right)$$

We focus on the $i^{\text{th}}$ dimension of the parameter vector $\theta_t \in R^d$. Subtract the scalar $\theta^*_{,i}$ and square both sides of the above update rule, we have,

$$(\theta_{t+1,i} - \theta^*_{,i})^2 = (\theta_{t,i} - \theta^*_{,i})^2 - \frac{2\alpha_t}{1-\beta_1^t} (\frac{\beta_{1,t}}{\sqrt{\widehat{v}_{t,i}}} m_{t-1,i} + (1 - \frac{\beta_{1,t}}{\sqrt{\widehat{v}_{t,i}}} g_{t,i})(\theta_{t,i} - \theta^*_{,i}) + \alpha_t^2 (\frac{\widehat{m}_{t,i}}{\sqrt{\widehat{v}_{t,i}}})^2$$

We can rearrange the above equation and use Young's inequality, $ab \leq a^2/2 + b^2/2$. Also, it can be shown that $\sqrt{\widehat{v}_{t,i}} = \sqrt{\sum_{j=1}^t (1-\beta_2)\beta_2^{t-j} g_{j,i}^2}/\sqrt{1-\beta_2^t} \leq \|g_{1:t,i}\|_2$ and $\beta_{1,t} \leq \beta_1$. Then

$$g_{t,i}(\theta_{t,i} - \theta^*_{,i}) = \frac{(1-\beta_1^t)\sqrt{\widehat{v}_{t,i}}}{2\alpha_t (1-\beta_{1,t})} \left( (\theta_{t,i} - \theta^*_{,t})^2 - (\theta_{t+1,i} - \theta^*_{,i})^2 \right)$$
$$+ \frac{\beta_{1,t}}{(1-\beta_{1,t})} \frac{\widehat{v}_{t-1,i}^{\frac{1}{4}}}{\sqrt{\alpha_{t-1}}} (\theta^*_{,i} - \theta_{t,i}) \sqrt{\alpha_{t-1}} \frac{m_{t-1,i}}{\widehat{v}_{t-1,i}^{\frac{1}{4}}} + \frac{\alpha_t(1-\beta_1^t)\sqrt{\widehat{v}_{t,i}}}{2(1-\beta_{1,t})} (\frac{\widehat{m}_{t,i}}{\sqrt{\widehat{v}_{t,i}}})^2$$
$$\leq \frac{1}{2\alpha_t(1-\beta_1)} \left( (\theta_{t,i} - \theta^*_{,t})^2 - (\theta_{t+1,i} - \theta^*_{,i})^2 \right) \sqrt{\widehat{v}_{t,i}} + \frac{\beta_{1,t}}{2\alpha_{t-1}(1-\beta_{1,t})} (\theta^*_{,i} - \theta_{t,i})^2 \sqrt{\widehat{v}_{t-1,i}}$$
$$+ \frac{\beta_1 \alpha_{t-1}}{2(1-\beta_1)} \frac{m_{t-1,i}^2}{\sqrt{\widehat{v}_{t-1,i}}} + \frac{\alpha_t}{2(1-\beta_1)} \frac{\widehat{m}_{t,i}^2}{\sqrt{\widehat{v}_{t,i}}}$$

We apply Lemma 10.4 to the above inequality and derive the regret bound by summing across all the dimensions for $i \in 1, ..., d$ in the upper bound of $f_t(\theta_t) - f_t(\theta^*)$ and the sequence of convex functions for $t \in 1, ..., T$:

$$R(T) \leq \sum_{i=1}^d \frac{1}{2\alpha_1(1-\beta_1)} (\theta_{1,i} - \theta^*_{,i})^2 \sqrt{\widehat{v}_{1,i}} + \sum_{i=1}^d \sum_{t=2}^T \frac{1}{2(1-\beta_1)} (\theta_{t,i} - \theta^*_{,i})^2 (\frac{\sqrt{\widehat{v}_{t,i}}}{\alpha_t} - \frac{\sqrt{\widehat{v}_{t-1,i}}}{\alpha_{t-1}})$$
$$+ \frac{\beta_1 \alpha G_\infty}{(1-\beta_1)\sqrt{1-\beta_2}(1-\gamma)^2} \sum_{i=1}^d \|g_{1:T,i}\|_2 + \frac{\alpha G_\infty}{(1-\beta_1)\sqrt{1-\beta_2}(1-\gamma)^2} \sum_{i=1}^d \|g_{1:T,i}\|_2$$
$$+ \sum_{i=1}^d \sum_{t=1}^T \frac{\beta_{1,t}}{2\alpha_t(1-\beta_{1,t})} (\theta^*_{,i} - \theta_{t,i})^2 \sqrt{\widehat{v}_{t,i}}$$

From the assumption, $\|\theta_t - \theta^*\|_2 \leq D$, $\|\theta_m - \theta_n\|_\infty \leq D_\infty$, we have:

$$R(T) \leq \frac{D^2}{2\alpha(1-\beta_1)}\sum_{i=1}^d \sqrt{T\widehat{v}_{T,i}} + \frac{\alpha(1+\beta_1)G_\infty}{(1-\beta_1)\sqrt{1-\beta_2}(1-\gamma)^2}\sum_{i=1}^d \|g_{1:T,i}\|_2 + \frac{D_\infty^2}{2\alpha}\sum_{i=1}^d\sum_{t=1}^t \frac{\beta_{1,t}}{(1-\beta_{1,t})}\sqrt{t\widehat{v}_{t,i}}$$

$$\leq \frac{D^2}{2\alpha(1-\beta_1)}\sum_{i=1}^d \sqrt{T\widehat{v}_{T,i}} + \frac{\alpha(1+\beta_1)G_\infty}{(1-\beta_1)\sqrt{1-\beta_2}(1-\gamma)^2}\sum_{i=1}^d \|g_{1:T,i}\|_2$$
$$+ \frac{D_\infty^2 G_\infty\sqrt{1-\beta_2}}{2\alpha}\sum_{i=1}^d\sum_{t=1}^t \frac{\beta_{1,t}}{(1-\beta_{1,t})}\sqrt{t}$$

We can use arithmetic geometric series upper bound for the last term:

$$\sum_{t=1}^t \frac{\beta_{1,t}}{(1-\beta_{1,t})}\sqrt{t} \leq \sum_{t=1}^t \frac{1}{(1-\beta_1)}\lambda^{t-1}\sqrt{t}$$
$$\leq \sum_{t=1}^t \frac{1}{(1-\beta_1)}\lambda^{t-1} t$$
$$\leq \frac{1}{(1-\beta_1)(1-\lambda)^2}$$

Therefore, we have the following regret bound:

$$R(T) \leq \frac{D^2}{2\alpha(1-\beta_1)}\sum_{i=1}^d \sqrt{T\widehat{v}_{T,i}} + \frac{\alpha(1+\beta_1)G_\infty}{(1-\beta_1)\sqrt{1-\beta_2}(1-\gamma)^2}\sum_{i=1}^d \|g_{1:T,i}\|_2 + \sum_{i=1}^d \frac{D_\infty^2 G_\infty\sqrt{1-\beta_2}}{2\alpha\beta_1(1-\lambda)^2}$$

$\square$